\documentclass[10pt,conference]{article}
\usepackage{spconf,amsmath,amsfonts,graphicx,caption}
\usepackage[hyphens]{url}
\usepackage{cite}
\usepackage[T1]{fontenc}
\usepackage{indentfirst}
\usepackage{multirow}
\usepackage{color}

\usepackage[hidelinks,bookmarks=false]{hyperref}

\title{A pipeline for lenslet light field quality enhancement}

\name{Pierre Matysiak, Mair\'ead Grogan, Mika\"el Le Pendu, Martin Alain, Aljosa Smolic}
\address{V-SENSE Project, School of Computer Science and Statistics, Trinity College, Dublin
\thanks{This publication has emanated from research conducted with the financial support of Science Foundation Ireland (SFI) under the Grant Number 15/RP/2776.
This work was supported by TCHPC (Research IT, Trinity College Dublin).
All calculations were performed on the Lonsdale cluster maintained by the Trinity Centre for High Performance Computing.
This cluster was funded through grants from SFI.}}%

\begin{document}
\fontsize{9.5}{11}\selectfont
\setlength{\belowdisplayskip}{0pt}
\setlength{\belowdisplayshortskip}{0pt}
\setlength{\abovedisplayskip}{0pt}
\setlength{\abovedisplayshortskip}{0pt}

\maketitle

\begin{abstract}
\small{\textbf{In recent years, light fields have become a major research topic and their applications span across the entire spectrum of classical image processing. Among the different methods used to capture a light field are the lenslet cameras, such as those developed by Lytro. While these cameras give a lot of freedom to the user, they also create light field views that suffer from a number of artefacts. As a result, it is common to ignore a significant subset of these views when doing high-level light field processing. We propose a pipeline to process light field views, first with an enhanced processing of RAW images to extract sub-aperture images, then a colour correction process using a recent colour transfer algorithm, and finally a denoising process using a state of the art light field denoising approach. We show that our method improves the light field quality on many levels, by reducing ghosting artefacts and noise, as well as retrieving more accurate and homogeneous colours across the sub-aperture images.
}}
\end{abstract}


\vspace{-10pt}
\section{Introduction}
\label{sec:intro}
\vspace{-5pt}

Light fields aim to capture all light rays passing through a given amount of the 3D space \cite{Levoy1996}.
Compared to traditional images representing a projection of light rays on a 2D plane, a 4D light field also contains the angular direction of the rays.
The light field of a real scene can be captured with different devices such as a single camera on a moving gantry, an array of cameras, or a plenoptic camera including an array of micro-lenses in front of its sensor. The latter has received a lot of attention since the commercialisation by the Lytro 
company of two successive models capable of capturing light fields with a dense angular sampling.
A concurrent plenoptic camera design called plenoptic 2.0 has also been proposed in \cite{Pleno2_0}. Unlike the former design (referred to as unfocused), each micro-lens produces a focused micro-image.

Because of the micro-lens array, the generation of exploitable images from the RAW sensor data is significantly more complex than with traditional cameras. Furthermore, there is no consensus on the light field representation to adopt. While plenoptic 2.0 cameras are generally used to directly render images at varying focus, the unfocused plenoptic cameras are better suited for the extraction of sub-aperture images (SAI) with a very wide depth of field, each corresponding to a viewpoint of the scene. Since the latter representation is more commonly used for various light field applications (e.g. depth estimation, compression),  we consider in our analysis the extraction of SAIs from unfocused plenoptic camera RAW data.
Among the different methods proposed for this task \cite{DansereauPW13, KAIST_Bary, Xu14Demultiplex,Seifi14Demosaic}, the most complete pipeline was developed by Dansereau et al. \cite{DansereauPW13}. It has been widely adopted by the light field research community. For example, it has a central role in the standardisation effort for light field compression as it is used as part of the JPEG PLENO \cite{JPEGPLENO} test set.
However, the extracted views may suffer from many artefacts including noise, unnatural horizontal stripes, ghosting effects on external SAIs, colour and brightness inconsistencies between SAIs, inaccurate colour balance, and important loss of dynamic range.
These limitations have a negative impact on most light field applications such as depth estimation, segmentation, rendering or compression (see \cite{LFOverview} for a comprehensive overview). Furthermore, external SAIs containing essential depth information are often ignored because of extreme distortions.
Note that, although the proprietary Lytro Desktop software overcomes many of these issues, it essentially targets the rendering of refocused images, and it is not suitable for generating an SAI array.

Later research on the subject in \cite{David2017,Seifi14Demosaic,Xu14Demultiplex,Lian16Demosaic,Yu12Demosaic,Huang14DictDemosaic} has essentially focused on adapting the demosaicing step which retrieves the RGB colour components of each pixel from the partial colour information actually captured by camera sensors. We believe that a more global analysis of the pipeline is also necessary.

In this paper, we present an enhanced processing pipeline for lenslet-based plenoptic cameras.
We first propose improvements within the RAW processing of \cite{DansereauPW13}. In particular, we show how the devignetting step (i.e. correction of lenslet vignetting) negatively impacts the overall image aspect (colour balance, brightness, loss of dynamic range), and how to correct it. We additionally recommend the use of white image guided interpolations \cite{David2017} to reduce the ghosting effect of external SAIs. Based on the observation that brightness and colour inconsistencies between SAIs can hardly be corrected in the early stages of the RAW processing without introducing other artefacts, we perform a post processing colour correction step. We finally analyse the noise level at each step of the process and suggest that additional denoising should be applied preferably after the colour correction.

\begin{figure}
\centering
\begin{minipage}[h]{.325\linewidth}
	\centerline{\includegraphics[width=\linewidth]{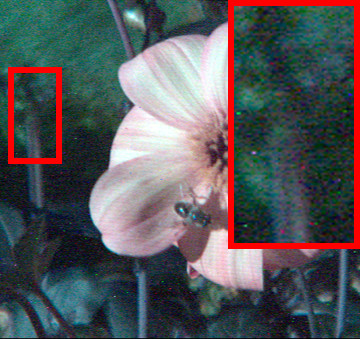}}
	\vspace{-4pt}
	\centerline{\small{(a)}}
\end{minipage}
\begin{minipage}[h]{.325\linewidth}
	\centerline{\includegraphics[width=\linewidth]{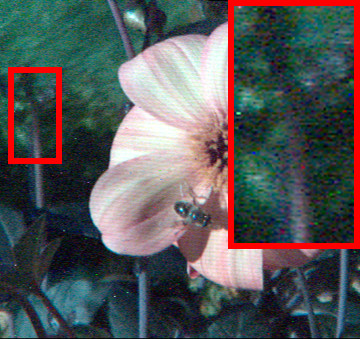}}
	\vspace{-4pt}
	\centerline{\small{(b)}}
\end{minipage}
\begin{minipage}[h]{.325\linewidth}
	\centerline{\includegraphics[width=\linewidth]{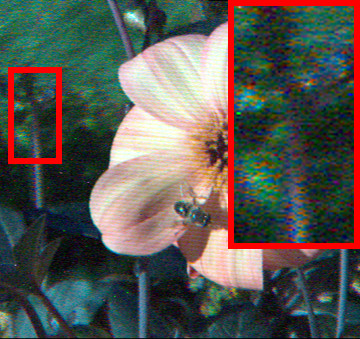}}
	\vspace{-4pt}
	\centerline{\small{(c)}}
\end{minipage}
\vspace{-9pt}
\caption{{\small Advantages and limitations of the White Image (WI) guided method of \cite{David2017}: (a) standard demosaicing \cite{Malvar04} and bicubic interpolations, (b) standard demosaicing \cite{Malvar04} and WI-guided interpolations, (c) WI-guided demosaicing and interpolations.}}
\label{fig:DemosaicInterp}
\vspace{-15pt}
\end{figure}

\vspace{-10pt}
\section{Proposed pipeline}
\label{sec:pipeline}
\vspace{-8pt}

\subsection{RAW Light Field Decoding}
\label{ssec:RAWDecoding}
\vspace{-7pt}
In this section, we analyse the early steps of the RAW processing : the lenslet devignetting, the demosaicing, and the interpolations required for compensating misalignments between the micro-lens array and the sensor grid.

\vspace{-8pt}
\subsubsection{Lenslet Devignetting}
\vspace{-5pt}
\label{sssec:Devignetting}
In \cite{DansereauPW13}, lenset devignetting is performed first as it results in more uniform brightness over the sensor array and thus, easier demosaicing. This step simply consists of a pixel-wise division of the RAW image by
a RAW White Image (WI) that exhibits the pattern of micro-lens vignetting.
Since the WI was previously taken by the same device as the picture to process, this division step does not only remove the vignetting pattern, but also implicitly normalises the colour responses of the red, green, and blue pixels on the sensor. However, the normalisation is not taken into account by traditional RAW processing \cite{DansereauPW13}. In particular, the white balance parameters, determined by the camera during the capture (either automatically or with user interaction) and stored as metadata, do not apply to RAW data with normalised RGB responses. In order to obtain the intended colour balance, we multiply the red and blue pixels of the WI by normalisation factors provided as metadata of the camera. Note that these factors may also be obtained by colour calibration of the sensor.

Since the pixel values of the WI are lower than 1 even at micro-lens centres, the devignetting in \cite{DansereauPW13} also increases the overall brightness of the light field. Bright areas reaching higher values than 1 after devignetting are considered as saturated in the rest of the process, and the information is lost. Therefore, we also apply a global normalisation of the WI by dividing all the pixels by its $99.9$\textsuperscript{th} percentile (we do not use the maximum value to exclude hot pixels).

\vspace{-5pt}
\subsubsection{Demosaicing and Interpolations}
\vspace{-5pt}
Previous analysis by David et al. \cite{David2017} have shown how standard demosaicing and interpolations introduced both ghosting artefacts and fading of the colours for the external SAIs. In order to reduce the problem, they adapted those steps by weighing the contribution of each pixel using the vignetting pattern of the White Image. Two observations can be made from their results. Firstly, the ghosting effect is essentially reduced by the adaptation of the interpolation step (see Fig. \ref{fig:DemosaicInterp}(b)). Secondly, while the modified demosaicing improves the overall colour consistency between SAIs, it may also create colour noise (see Fig. \ref{fig:DemosaicInterp}(c)).
Hence, we suggest that only the WI-guided interpolations should be used, and we propose in the next section a post-processing step to enforce colour homogeneity in the light field.


\vspace{-5pt}
\subsection{Colour Correction}
\vspace{-5pt}
To obtain homogeneous colours in the light field, we successively process each SAI by performing colour transfer from a reference SAI designated as palette image. Several propagation schemes defining the palette image for each SAI and the processing order are proposed in section \ref{sssec:Ccorr_prop}.

For a given pair of target and palette SAIs, our colour correction addresses the viewpoint disparity by incorporating correspondence estimations in the recent colour transfer algorithm of Grogan et al. \cite{Grogan15, Grogan2017, Grogan17} described in section \ref{sssec:colourTransfer}. Their global approach was found to be more robust to erroneous correspondences and outperforms existing colour correction methods \cite{Hwang2014, Bonneel2015, PITIE2007}. In our implementation, coarse-to-fine patch matching (CPM) \cite{Hu2016} was chosen to estimate pixel correspondences between the views since it is both accurate and efficient, and has been successfully used as an initialisation step for optical flow computation for light fields by Chen et al. \cite{Chen2017}. Once the pixel correspondences are estimated, we pass their colour values to the colour transfer algorithm.

\vspace{-5pt}
\subsubsection{Colour Transfer}
\vspace{-5pt}
\label{sssec:colourTransfer}
Given a set of $n$  colour correspondences ${(c_t^{(k)}, c_p^{(k)})}_{k=1\dots n}$ between the target and palette images, where the set of colours $c_t^{(k)}$ from the target image should correspond to the colours $c_p^{(k)}$ from the palette after recolouring, Grogan et al. \cite{Grogan2017,Grogan17} propose to fit a Gaussian Mixture Model to each set of correspondences as follows:
\begin{align}
p_t(x|\theta) = \sum_{k=1}^{\mathrm n} \frac{1}{\mathrm n} \mathcal{N}(x;\,\phi(c_t^{(k)}, \theta),\,h^2{\mathrm I})
\label{form:ptxtheta}
\end{align}
and
\begin{align}
p_p(x) = \sum_{k=1}^{\mathrm n} \frac{1}{\mathrm n} \mathcal{N}(x;\,c_p^{(k)},\,h^2{\mathrm I})
\label{form:ppx}
\end{align}

The vector $x\in\mathbb{R}^3$ takes values from a 3D colour space, and each Gaussian is associated with an identical isotropic covariance matrix, $h^2{\mathrm I}$. The colours $ \phi(c_t^{(k)}, \theta)$ are obtained by transforming ${c_t^{(k)}}$ by some transformation $\phi$ which depends on $\theta$. The goal is then to estimate the transformation $\phi$ that registers $p_t(x|\theta)$ to $p_p(x)$, and thus transforms the colour distribution of the target image to match that of the palette image. Grogan et al. propose letting $\phi$ be a global parametric thin plate spline transformation, and estimate the parameter $\theta$ controlling $\phi$ by minimising :
\begin{align}
\mathcal{C}(\theta) = - \langle p_t|p_p \rangle = \sum_{k=1}^n\,\frac{1}{n^2}\,\mathcal{N}(0;\,\phi(c_t^{(k)},\,\theta) - c_p^{(k)},\,2h^2{\mathrm I})
\label{form:optim_fun}
\end{align}

For our application, we found that using the LAB space representation of colours gave better results than RGB. We also altered the simulated annealing parameters proposed in \cite{Grogan2017} to ensure that local minima were avoided during optimisation.

\begin{figure*}[t]
	\footnotesize
	\centering
	\begin{tabular}{ccc}
	\includegraphics[width=0.3\linewidth,clip]{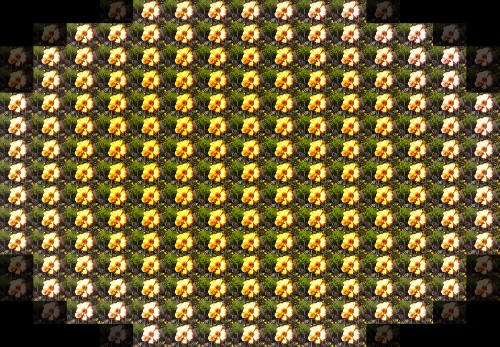} &
	\includegraphics[width=0.3\linewidth,clip]{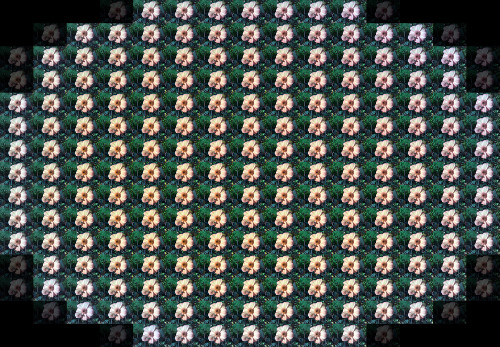} & 
	\includegraphics[width=0.3\linewidth,clip]{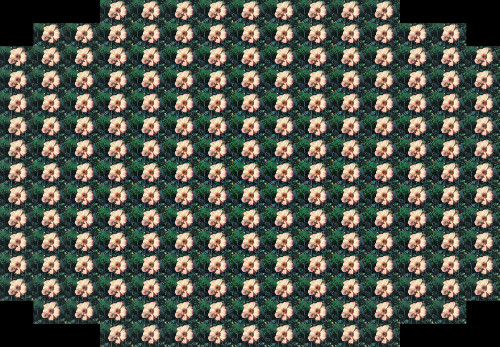} \\
	(a) Original RAW decoding \cite{DansereauPW13} &
	(b) Proposed RAW decoding &
	(c) Recoloured \\
	\end{tabular}
	\vspace{-8pt}
	\caption{\small{Overview of all sub-aperture images in a light field. Showcases the differences in appearance between the original and proposed RAW decoding, and the homogeneity of the colours across views after colour correction. }}
	\label{fig:LF_recolor_example}
	\vspace{-15pt}
\end{figure*}

\vspace{-5pt}
\subsubsection{Propagation}
\vspace{-5pt}
\label{sssec:Ccorr_prop}
To guarantee colour homogeneity over the whole light field, we investigate three propagation schemes. The first is a more naive approach in which we recolour every SAI in the light field taking the centre view as palette, as it has the most accurate colours. This involves computing correspondences between each SAI and the centre one and passing them to Eqs. \eqref{form:ptxtheta} and \eqref{form:ppx} to perform colour correction. While these correspondences are accurate in many cases, as the disparity increases over the light field, fewer accurate correspondences are available which can affect the quality of the recoloured external views.

To combat this, we investigated a second scheme which involves propagating the colours incrementally starting from the centre view, along the centre column, and then along each row. This is achieved by first taking the centre view as palette and using it to recolour its two outer neighbouring views in the column. Once corrected, these two views are used to recolour their outer neighbouring views. This process is repeated until all views in the column are corrected. Following a similar process, the colours from the centre column are propagated out to each of the rows, with the centre view in each row initialised as palette. This scheme ensures that neighbouring views have very few colour differences between them.

The final scheme is a combination of the previous two, with each view (apart from the centre one) recoloured using the colour correspondences from its already corrected inner column or row neighbour as well as those from the centre view, and combining them when passing them to Eqs. \eqref{form:ptxtheta} and \eqref{form:ppx}, in order to maintain colour consistency across the light field. The three methods will henceforth be referred to as `centre', `prop' and `prop+centre'.

\vspace{-5pt}
\subsection{Denoising}
\vspace{-5pt}

In addition to the colour artefacts corrected by the previous step of the proposed pipeline, light fields captured with the Lytro Illum are known to exhibit camera noise pattern.
We thus propose to apply denoising as a final step.
For that purpose, we use the state-of-the art LFBM5D filter introduced in \cite{Alain2017}.
This filter takes full advantage of the 4D nature of light fields by creating disparity compensated 4D patches which are then stacked together with similar 4D patches along a 5\textsuperscript{th} dimension.
These 5D patches are then filtered in the 5D transform domain, obtained by cascading a 2D spatial transform, a 2D angular transform, and a 1D transform applied along the similarities.

Denoising is therefore applied after colour correction, so that the dark corner SAIs can be denoised together with the rest of the light field.


\vspace{-10pt}
\section{Results}
\label{sec:results}
\vspace{-7pt}

We present results of our pipeline applied on a subset of the freely available EPFL \cite{Rerabek2016} and INRIA \cite{LePendu2018} datasets captured with Lytro Illum cameras. Detailed dataset composition and additional results are also available online\footnote{\url{https://v-sense.scss.tcd.ie/?p=1548}}.

\begin{figure}[b]
\vspace{-10pt}
\centering
\begin{minipage}[h]{.4\linewidth}
	\centerline{\includegraphics[width=\linewidth]{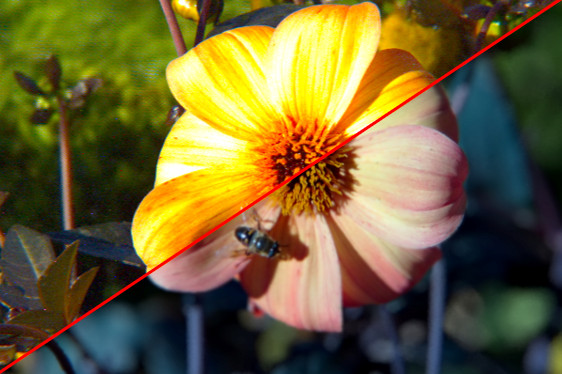}}
	\vspace{-4pt}
	\centerline{\footnotesize{(a) Method \cite{DansereauPW13} vs Lytro Desktop}}
\end{minipage}
\hspace{20pt}
\begin{minipage}[h]{.4\linewidth}
	\centerline{\includegraphics[width=\linewidth]{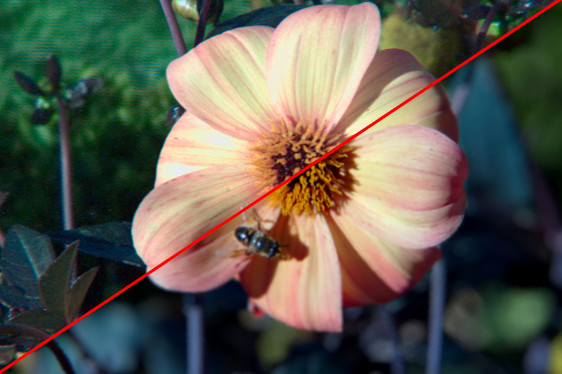}}
	\vspace{-4pt}
	\centerline{\footnotesize{(b) Our method vs Lytro Desktop}}
\end{minipage}
\vspace{-7pt}
\caption{{\small Above red line: central SAI of the light field obtained with (a) Dansereau et al. \cite{DansereauPW13}, (b) Our method. Below red line: refocused image from Lytro Desktop proprietary software (using `as shot' white balance option).}}
\label{fig:OverallColors}
\vspace{-10pt}
\end{figure}

\vspace{-5pt}
\subsection{Colour Quality}
\vspace{-5pt}

We first show in Fig. \ref{fig:OverallColors} the importance of the simple normalisation steps proposed in section \ref{sssec:Devignetting} for the colour balance and overall brightness. For reference, the bottom right part of each sub-figure shows a refocused image obtained by the Lytro proprietary software with the intended colours (i.e. as displayed by the camera when taking the picture).
Note that the results of \cite{DansereauPW13} are often wrongly assumed to be gamma corrected, leading to exaggerated contrasts and colour saturation. For a fair comparison, we performed standard sRGB gamma correction for both methods.

Regarding colour correction, we determined the best recolouring scheme (see Sec. \ref{sssec:Ccorr_prop}) by comparing their results using a number of metrics : S-CIELab \cite{SCIE1997}, a global histogram distance, and a blind noise level estimation \cite{Liu2013}. We investigated other metrics for comparison (PSNR, SSIM \cite{ SSIMWang04}, CID \cite{Lissner13}),  but found that the three presented here were the most comprehensive. To determine the accuracy of the colour correction with respect to the centre view, we computed the difference between the centre SAI and each SAI in the light field 
using S-CIELab and the histogram distance, and averaged the results over SAIs. As S-CIELab compares local colour differences between images, the disparity between SAIs may affect the evaluation. However, as all methods are compared on the same set of light fields with the same disparity differences, values are still indicative of colour correction accuracy.  The global histogram metric on the other hand is more robust to these disparity changes. 
For a pair of images, the histogram distance is measured as the average $\chi$-square differences between their L, a*, and b* histograms, each computed on 25 bins.

\begin{figure}[t]
    \begin{center}
        \begin{tabular}{c}
            \includegraphics[width=0.7\linewidth]{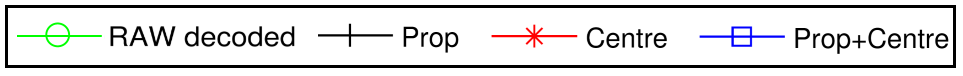} \\
            \includegraphics[width=0.48\linewidth]{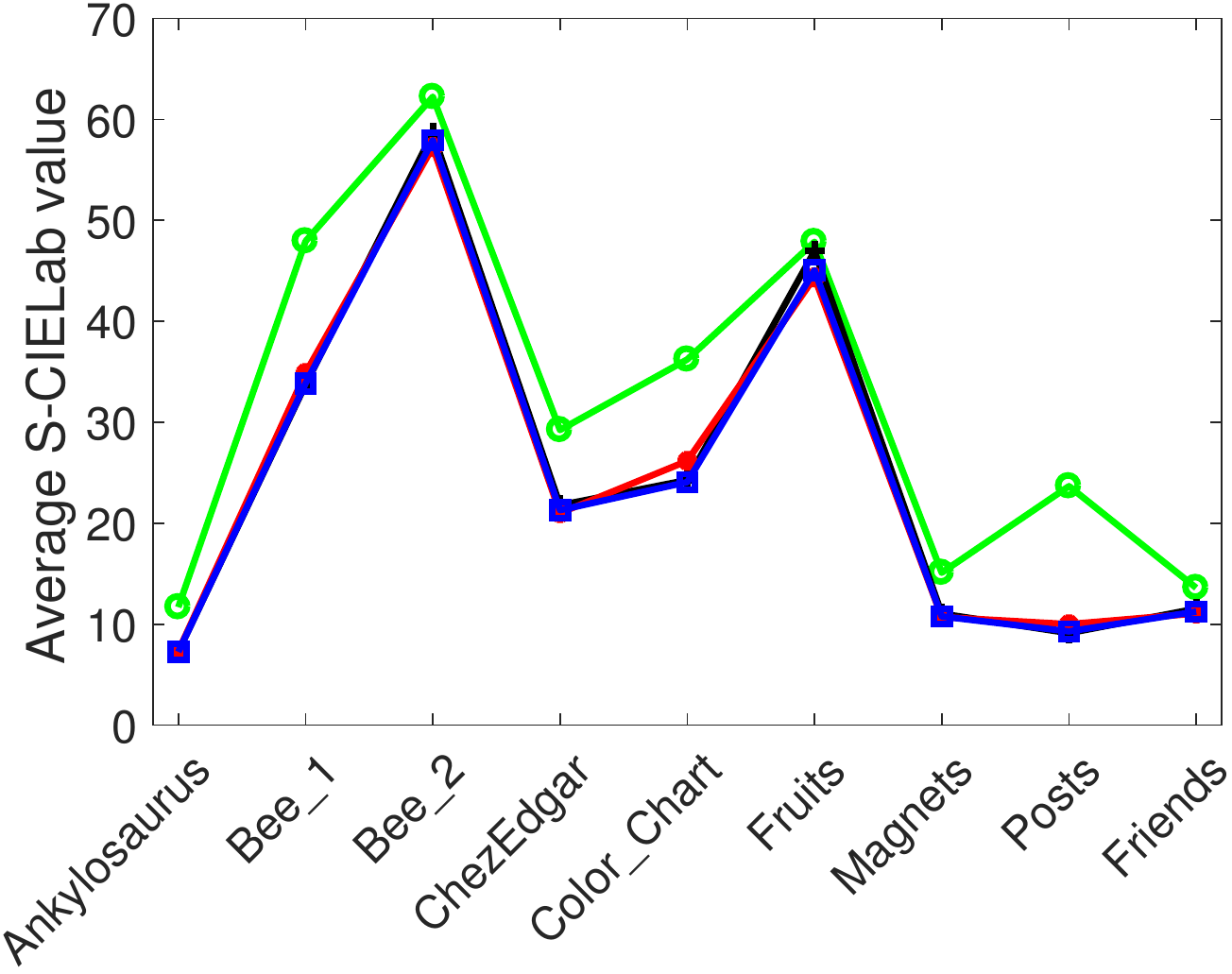}
            \includegraphics[width=0.48\linewidth]{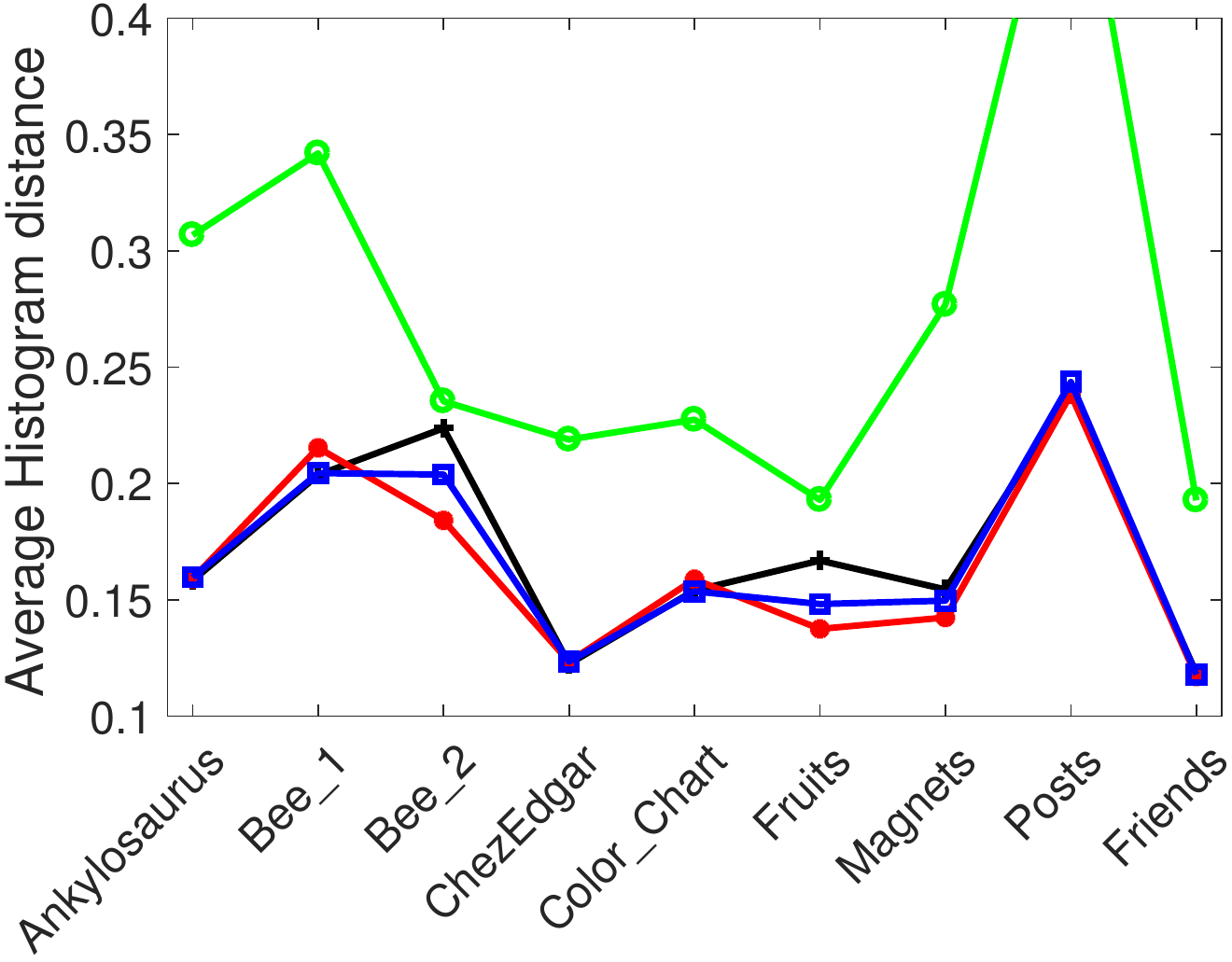} \\
        \end{tabular}
    \vspace{-10pt}
    \caption{\small{Metric comparison, using S-CIELab \cite{SCIE1997} and histogram distance. Lower values are better. It shows all three schemes are outputting comparable results.}}
    \label{fig:metrics}
    \end{center}
    \vspace{-15pt}
\end{figure}

\begin{figure}[t]
    \begin{center}
        \begin{tabular}{c}
            \includegraphics[width=0.8\linewidth]{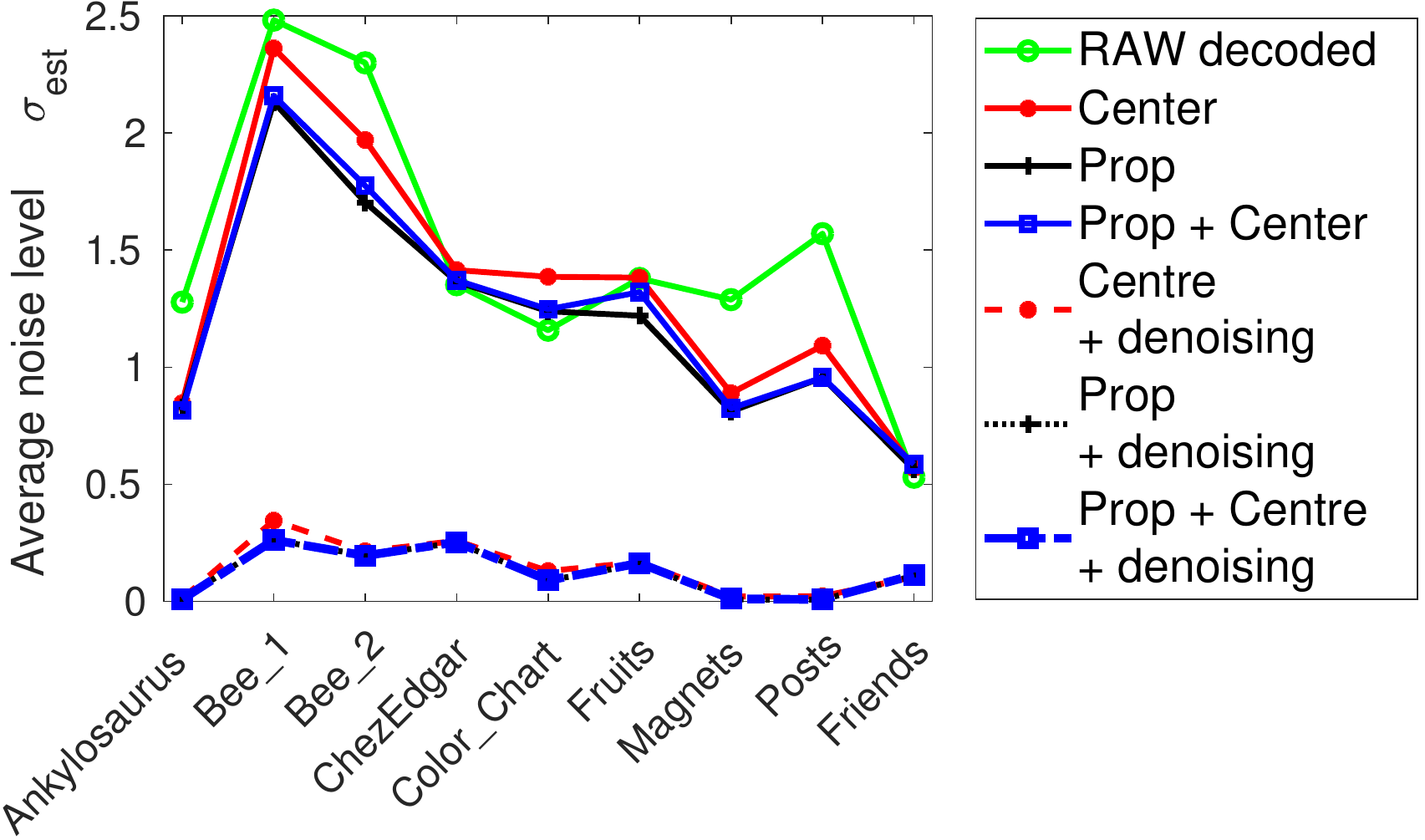} \\ 
        \end{tabular}
    \vspace{-10pt}
    \caption{\small{Noise level estimated for each light field using \cite{Liu2013}, before and after denoising.}}
    \label{fig:noise_lvl}
    \end{center}
    \vspace{-25pt}
\end{figure}

From Fig. \ref{fig:metrics} we can see that all approaches improve the overall colour similarity between the centre view and the remainder of the light field, with the `centre' scheme marginally capturing the colours of the centre view more faithfully according to S-CIELab and the histogram distance. However, as previously mentioned, it displays inaccuracies in some unfavourable cases, when the scene contains plain textureless colours and accurate correspondences are difficult to compute (e.g. Color\_Chart in Fig. \ref{fig:Reco_bee_cchart_rose}).
On the other hand, the `prop' and `prop+centre' schemes produce comparable results to `centre' with respect to S-CIELab and the histogram distance, and create less noise (Fig. \ref{fig:noise_lvl}). They also ensure that neighbouring SAIs have consistent colours. While we found that using the `prop+centre' scheme can sometimes give a more accurate colour correction compared to `prop', using additional correspondences increases the computational complexity of the colour correction step, and the decision to use these additional correspondences is left to the user.
\begingroup
\setlength{\tabcolsep}{2pt}
\begin{figure}
    \centering
    \begin{tabular}{@{}lccr}
        \includegraphics[width=0.235\linewidth,clip]{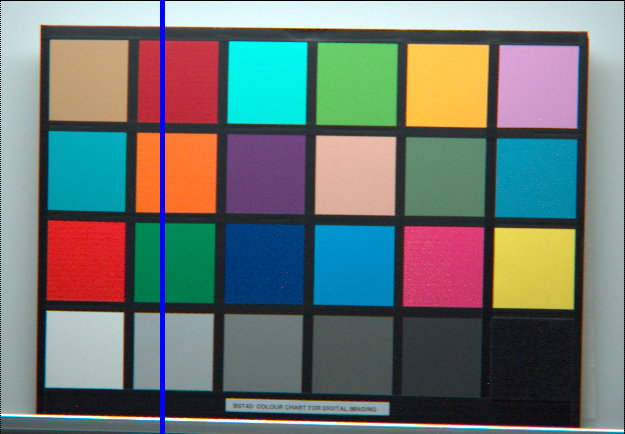} &
        \includegraphics[width=0.235\linewidth,clip]{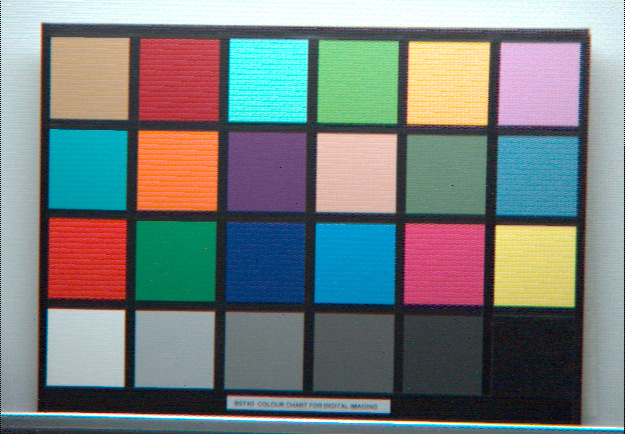} &
        \includegraphics[width=0.235\linewidth,clip]{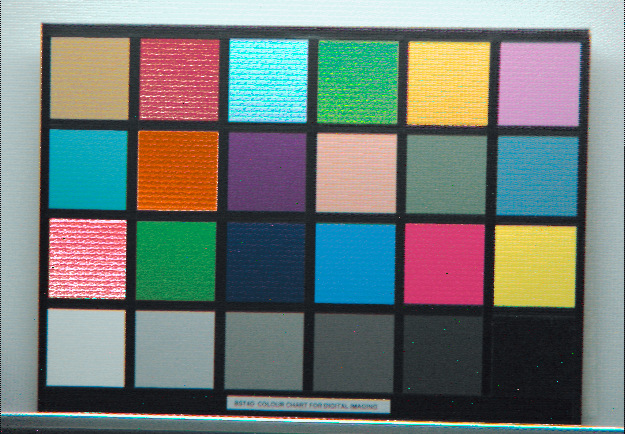} &
        \includegraphics[width=0.235\linewidth,clip]{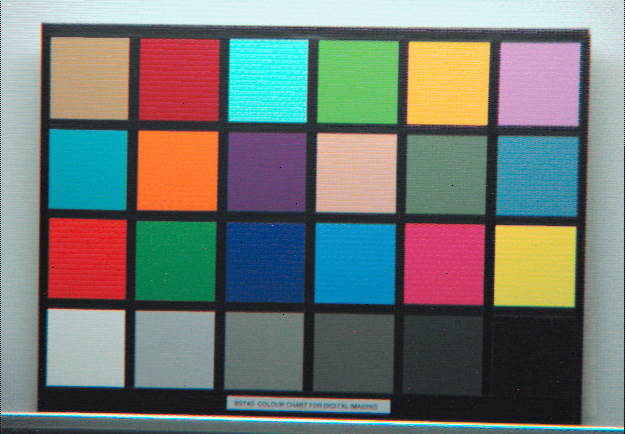} \\
        \includegraphics[width=0.235\linewidth,clip]{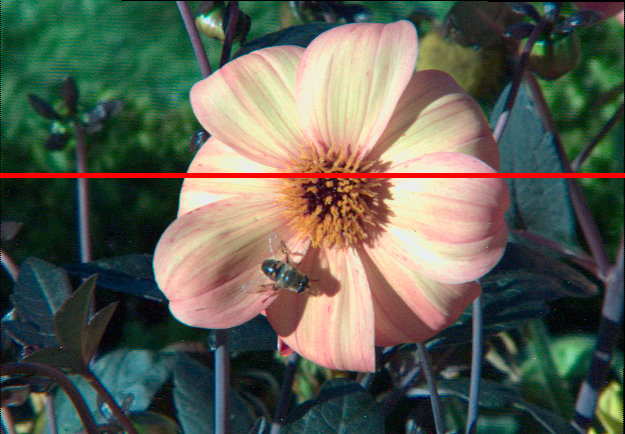} &
        \includegraphics[width=0.235\linewidth,clip]{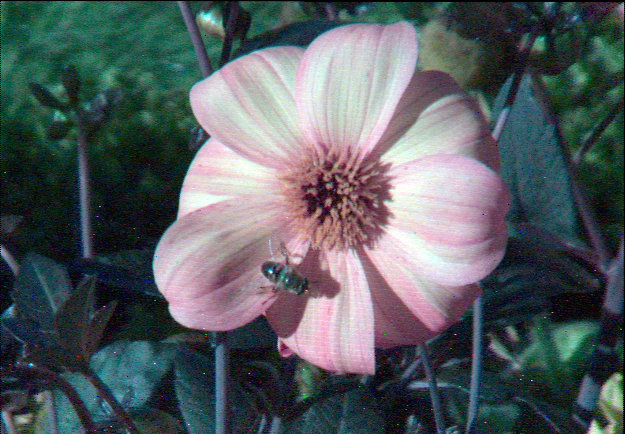} &
        \includegraphics[width=0.235\linewidth,clip]{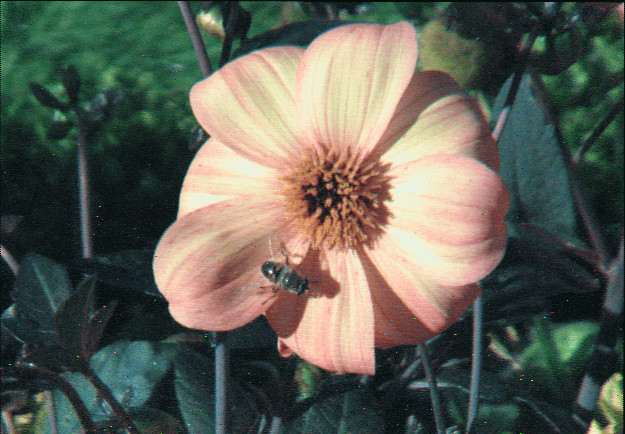} &
        \includegraphics[width=0.235\linewidth,clip]{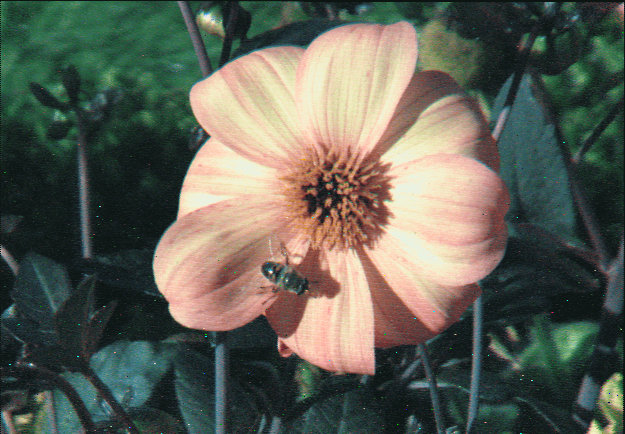}\\
    \end{tabular}
    \vspace{-10pt}
    \caption{\small{Recolouring examples. First column shows the centre SAI (red and blue lines are used to create the EPIs in Fig. \ref{fig:Epipolar}); second column is an SAI picked on the border of the light field; third and fourth columns are the same after recolouring using the `centre' and `prop+centre' schemes respectively.}}
    \label{fig:Reco_bee_cchart_rose}
    \vspace{-10pt}
\end{figure}
\endgroup

\begingroup
\setlength{\tabcolsep}{1pt}
\begin{figure}
    \centering
    \begin{tabular}{@{}cccc}
        \begin{minipage}[h]{0.46\linewidth}
            \begin{minipage}[h]{\linewidth}
                \includegraphics[width=\linewidth]{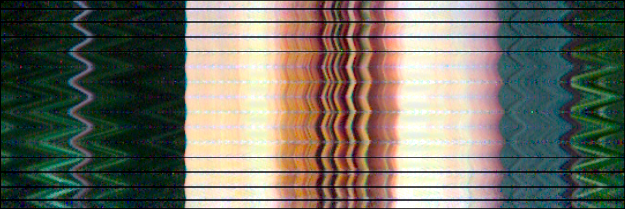}
                \vspace{0pt}\vspace{-17pt}
                \centerline{\small{(a)}}
            \end{minipage}
            \begin{minipage}[h]{\linewidth}
                \includegraphics[width=\linewidth]{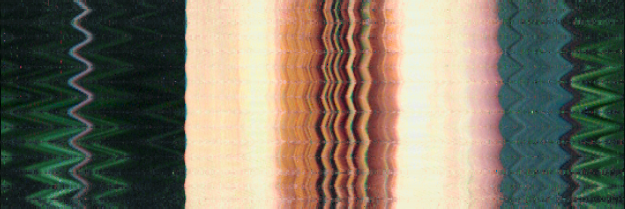}
                \vspace{0pt}\vspace{-17pt}
                \centerline{\small{(b)}}
            \end{minipage}
            \begin{minipage}[h]{\linewidth}
                \includegraphics[width=\linewidth]{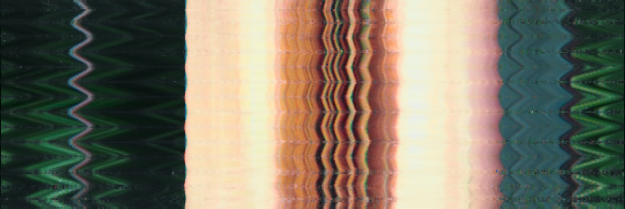}
                \vspace{0pt}\vspace{-15pt}
                \centerline{\small{(c)}}
            \end{minipage}
        \end{minipage} &

        \begin{minipage}[h]{0.17\linewidth}
            \includegraphics[width=\linewidth,height=3.17\linewidth]{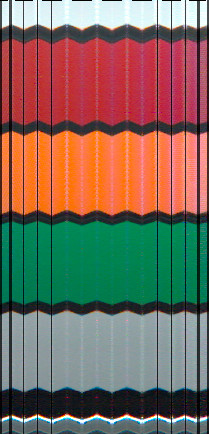}
            \vspace{0pt}\vspace{-15pt}
	        \centerline{\small{(d)}}
        \end{minipage} &
        \begin{minipage}[h]{0.17\linewidth}
            \includegraphics[width=\linewidth,height=3.17\linewidth]{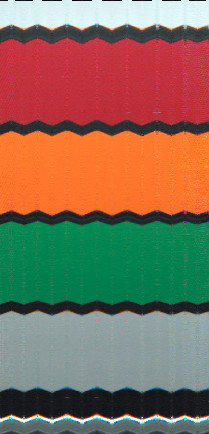}
            \vspace{0pt}\vspace{-15pt}
	        \centerline{\small{(e)}}
        \end{minipage} &
        \begin{minipage}[h]{0.17\linewidth}
            \includegraphics[width=\linewidth,height=3.17\linewidth]{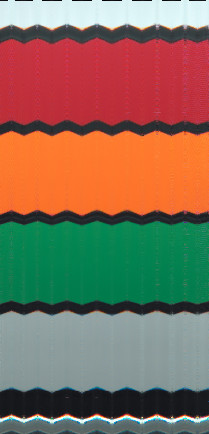}
            \vspace{0pt}\vspace{-15pt}
	        \centerline{\small{(f)}}
        \end{minipage}
    \end{tabular}
    \vspace{-10pt}
    \caption{\small{Stacked epipolar images showcasing colour differences in the LFs Bee\_2 (a,b,c) and Color\_Chart (d,e,f): after our RAW decoding (a,d), after `prop+centre' recolouring (b,e), and after denoising (c,f). Dark lines in (a,d) are caused by the dark SAIs in the corner of the light field (we only excluded the most extreme ones which are completely black and cannot be corrected). Selected lines are shown in Fig. \ref{fig:Reco_bee_cchart_rose}. \textbf{As with Fig. \ref{fig:Reco_bee_cchart_rose}, best viewed in colour and zoomed in.}}}
    \label{fig:Epipolar}
    \vspace{-15pt}
\end{figure}
\endgroup

We visually assess the results of our recolouring method in Figs. \ref{fig:LF_recolor_example}, \ref{fig:Reco_bee_cchart_rose} and \ref{fig:Epipolar}. 
The results are visually pleasing, with smooth transitions between consecutive views, seen in Fig. \ref{fig:LF_recolor_example}, and the colours overall remaining consistent with those in the centre view (see also Fig. \ref{fig:Reco_bee_cchart_rose}). This is particularly visible when computing epipolar images (as seen in Fig. \ref{fig:Epipolar}), which consist of stacks of the same horizontal or vertical line of pixels taken across all the views of the light field. These images show a clear improvement in keeping the colours consistent over the whole light field, which is further improved after the denoising process. Fig. \ref{fig:LF_recolor_example} shows that our colour correction also successfully recolours the dark corner images in the light field, which can then be taken advantage of by other processing tools.  


\vspace{-5pt}
\subsection{Noise Analysis}
\vspace{-5pt}
In order to quantify the noise reduction, we perform blind noise level estimation \cite{Liu2013} before and after denoising.
Assuming an Additive White Gaussian Noise (AWGN), we estimate the noise standard deviation $\sigma_{est}$ for each SAI, and then compute the average over the whole light field. 
Although the AWGN model is a simplification of the actual noise, our intent here is to provide a relative comparison of the different schemes rather than an absolute noise measure.
We report in Fig. \ref{fig:noise_lvl} the estimated values for each light field.
Results show that overall the colour correction step can slightly decrease the noise level.
However, we observe that in some cases the noise is amplified (e.g. Color\_Chart), which further justifies applying denoising last.
The noise level is clearly reduced for all approaches after applying the denoising.
As mentioned previously, the `centre' scheme exhibits a higher noise level than the other two tested approaches, even after denoising.
A visual comparison before and after denoising is shown in Fig. \ref{fig:Epipolar}.

\vspace{-5pt}
\section{Conclusion}
\label{sec:conclusion}
\vspace{-5pt}

We have presented a pipeline which aims at substantially improving upon the overall visual quality of SAIs in a light field. The final results show that every processing step provides  necessary and complementary benefits. We feel that providing a way to enhance the available lenslet camera datasets is necessary as such a complete approach does not currently exist. Visual inspection as well as metric comparison show that our method provides significant improvement on the quality of the light field views, counteracting the unfortunate side-effects lenslet cameras suffer from. Furthermore, by allowing the use of a higher number of views when performing high-level light field processing, such as depth estimation, segmentation or rendering, we hope to improve the results of those algorithms.

\bibliographystyle{ieee}

\end{document}